\documentclass[10pt,twocolumn,letterpaper]{article}

\usepackage{cvpr}              

\usepackage[ruled,vlined]{algorithm2e}

\definecolor{cvprblue}{rgb}{0.21,0.49,0.74}
\usepackage[pagebackref,breaklinks,colorlinks,allcolors=cvprblue]{hyperref}

\usepackage{booktabs}
\usepackage{caption}
\usepackage{multirow}

\def\paperID{*****} 
\def\confName{CVPR}
\def\confYear{2026}
\newcommand{\RLPOLICY}{PHASE}

\title{Heterogeneous Self-Play for Realistic Highway Traffic Simulation}

\author{Jinkai Qiu\\
PlusAI\\
{\tt\small jinkai.qiu@plus.ai}
\and
Alessandro Saviolo\\
PlusAI\\
{\tt\small alessandro.saviolo@plus.ai}
\and
Chaojie Wang\\
PlusAI\\
{\tt\small chaojie.wang@plus.ai}
\and
Mingke Wang\\
University of Michigan - Ann Arbor\\
{\tt\small mingkew@umich.edu}
\and
Xiaoyu Huang\\
PlusAI\\
{\tt\small xiaoyu.huang@plus.ai}
}

\begin{document}
\maketitle

\begin{abstract}
Realistic highway simulation is critical for scalable safety evaluation of autonomous vehicles, particularly for interactions that are too rare to study from logged data alone. Yet highway traffic generation remains challenging because it requires broad coverage across speeds and maneuvers, controllable generation of rare safety-critical scenarios, and behavioral credibility in multi-agent interactions. We present \textbf{\RLPOLICY}, \textbf{P}olicy for \textbf{H}eterogeneous \textbf{A}gent \textbf{S}elf-play on \textbf{E}xpressway, a context-aware self-play framework that addresses these three requirements through explicit per-agent conditioning for controllability, synthetic scenario generation for broad highway coverage, and closed-loop multi-agent training for realistic interaction dynamics. \textit{\RLPOLICY} further supports different vehicle profiles, for example, passenger cars and articulated trailer trucks, within a single policy via vehicle-aware dynamics and context-conditioned actions, and stabilizes self-play with early termination of unrecoverable states, at-fault collision attribution, highway-aware reward shaping, coupled curricula, and robust policy optimization. Despite being trained only on synthetic data, \textit{\RLPOLICY} transfers zero-shot to 512 unseen high-interaction real scenarios in exiD, achieving a 96.3\% success rate and reducing ADE/FDE from 6.57/12.07 m to 2.44/5.25 m relative to a prior self-play baseline. In a learned trajectory embedding space, it also improves behavioral realism over IDM, reducing Fr\'echet trajectory distance by 13.1\% and energy distance by 20.2\%. These results show that synthetic self-play can provide a scalable route to controllable and realistic highway scenario generation without direct imitation of expert logs.
\end{abstract}    

\section{Introduction}
Highway driving is one of the most safety-critical settings for autonomous vehicles: interactions unfold at high speed, small prediction errors can escalate quickly, and rare events such as aggressive cut-ins or dense merging are difficult to capture in sufficient quantity from real-world logs alone. This makes simulation a core tool for development and validation. A highway simulator, however, must do more than replay recorded traffic. It must expose safety-critical interactions at scale, remain stable in closed-loop rollouts, and support stress testing across diverse traffic regimes.

Existing approaches only partially satisfy these requirements. Log replay and supervised trajectory prediction preserve realistic behavior, but are limited by the behaviors present in the data and can drift under closed-loop execution~\cite{lasil, review}. Rule-based traffic models offer controllability, but often rely on simplified assumptions that fail to capture the richness of real multi-agent interactions. Recent self-play reinforcement learning (RL) methods offer a promising alternative by learning interactive behavior directly from closed-loop experience, and have shown strong results in urban driving~\cite{cornelisse2025buildingreliablesimdriving,gigaflow}, including on benchmarks derived from the Waymo Open Motion Dataset (WOMD)~\cite{WOMD}. Yet these methods have been developed primarily for urban regimes and do not directly address the demands of highway traffic.

Highway simulation poses a distinct challenge. Compared with urban driving, highway traffic spans a broader speed range, exhibits greater variation in density and interaction style, and includes stronger physical heterogeneity across vehicle classes. A practical highway simulator must therefore satisfy three requirements simultaneously: broad coverage across speeds and maneuvers, controllable generation of rare safety-critical scenarios, and realistic multi-agent behavior in closed-loop interaction.

\begin{figure*}[t]
  \centering
  \includegraphics[width=\textwidth]{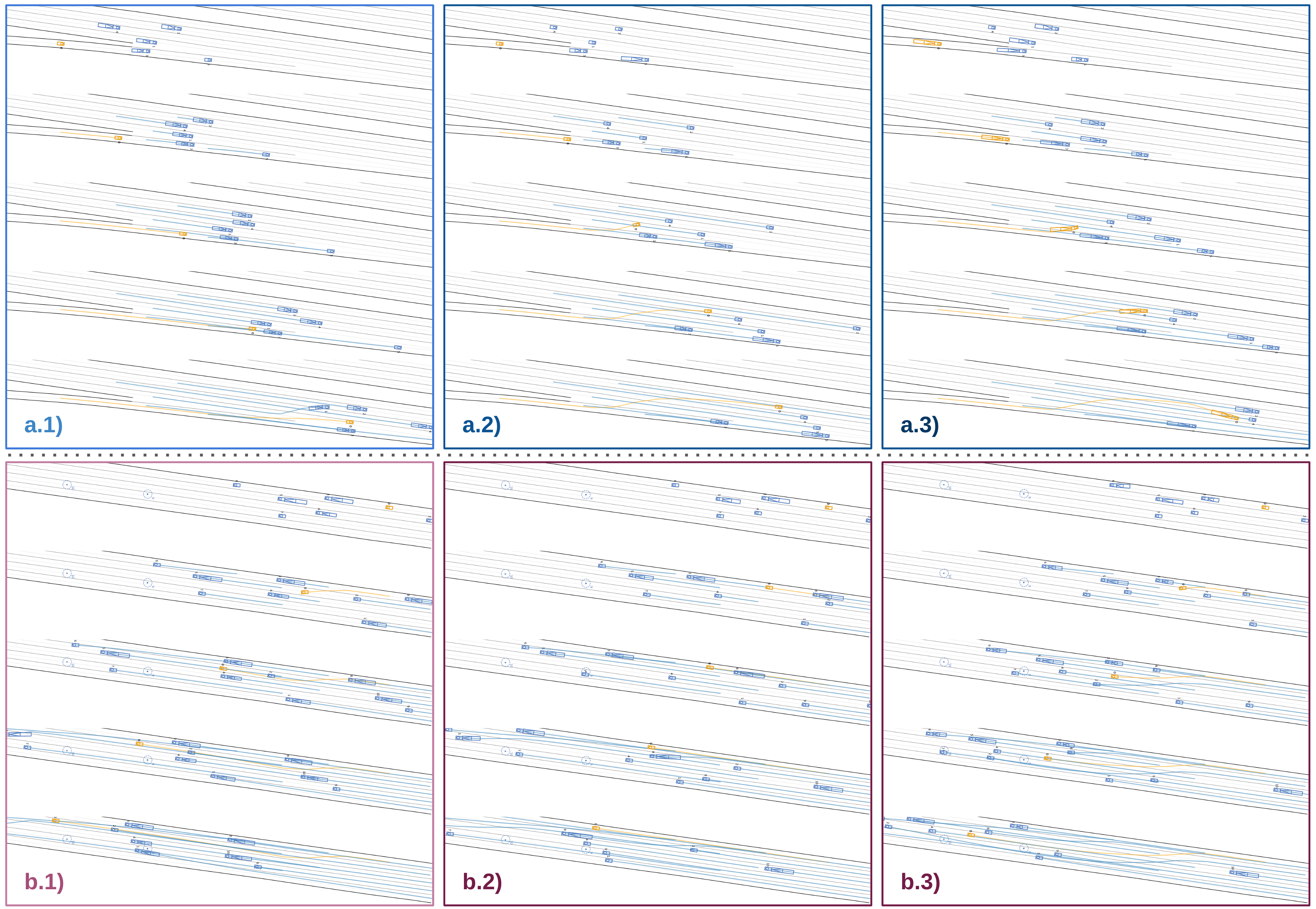}
  \caption{\textbf{Effect of context conditioning on closed-loop highway behavior.} The three panels in each row share the same initial scenario; only the context-conditioning variables $\mathcal{C}$ (Section~\ref{sec:taskdef}) are changed across columns. Despite identical initialization, the conditioned policy generates distinct closed-loop trajectory patterns within each triplet. Joint conditioning on $\mathcal{C}$ and each agent's Cartesian goal $g_i$ enables explicit control over scenario outcomes, as illustrated by vehicle~0 (orange), which alternates between overtaking and being overtaken.
  }
  \label{fig:diversity}
\end{figure*}

We address these challenges with \RLPOLICY, a self-play framework for controllable heterogeneous highway traffic generation. We formulate the problem as a conditioned goal-reaching Partially Observable Stochastic Game (POSG), in which each agent is assigned a Cartesian goal together with context variables such as target speed, longitudinal action range, and vehicle type. This conditioning provides explicit behavioral control, while a single policy handles both passenger cars and articulated tractor-trailers. As illustrated in Figure~\ref{fig:diversity}, varying the context variables while holding the base scene fixed produces distinct interaction patterns and allows users to steer scenario outcomes directly. To achieve broad coverage without relying on expert trajectory imitation, we train entirely on synthetic scenarios generated by an offline--online pipeline: lane-graph search constructs diverse start--goal pools offline, and online sampling randomizes agent count, lane-change composition, kinematics, and geometry. To preserve realistic behavior under self-play, we further combine simulation mechanisms, highway-aware reward design, coupled curricula, and robust policy optimization.

We evaluate \textit{\RLPOLICY} in two complementary settings. On exiD~\cite{exid}, \textit{\RLPOLICY} transfers zero-shot to unseen real highway scenarios and substantially outperforms a prior self-play baseline in both success rate and displacement error. In a learned trajectory embedding space built from proprietary real highway logs, it also produces trajectory distributions that align more closely with real traffic than IDM~\cite{idm}.

Our main contributions are as follows:
\begin{itemize}
  \item \textbf{A conditioned formulation for controllable heterogeneous highway simulation.} We formulate highway traffic generation as a conditioned goal-reaching POSG in which a single policy controls both passenger cars and articulated tractor-trailers across a 0--40~m/s operating range through joint goal and context conditioning.
  \item \textbf{A scalable synthetic scenario generation pipeline for broad highway coverage.} We introduce an offline--online generator that combines lane-graph endpoint search with controllable world sampling to produce diverse, map-consistent highway scenes without direct imitation from expert trajectory logs.
  \item \textbf{A stable self-play training recipe for highway traffic.} We combine simulation mechanisms, highway-aware reward shaping, coupled curricula, and robust optimization choices to make learning-from-scratch self-play practical in heterogeneous highway regimes.
  \item \textbf{Strong real-world transfer and improved behavioral realism.} \textit{\RLPOLICY} generalizes zero-shot to exiD and produces trajectory distributions that more closely match real highway driving than prior self-play and classical controller baselines.
\end{itemize}
\section{Related Work}
\label{sec:related}

\textbf{Self-play RL for driving and autonomy.}
Self-play RL has been effective in complex multi-agent settings, from board games \cite{Silver2017AlphaGoZero, Silver2018AlphaZero} to large-scale strategy environments \cite{OpenAI2019Dota2}. Recently, this paradigm has been adapted to autonomous driving, where interaction is inherently multi-agent and closed-loop. Methods such as \cite{cornelisse2025buildingreliablesimdriving, gigaflow} show that large-scale self-play can learn robust driving policies in urban environments. However, existing driving applications have focused primarily on lower-speed urban regimes, where the dominant challenges are negotiation and topological complexity at intersections. In contrast, we study high-speed highway traffic, where smaller control errors can have larger consequences and where stable self-play must handle broader speed ranges and stronger vehicle heterogeneity.

\textbf{Large-scale driving simulators.}
GPU-accelerated simulators have made large-scale training for interactive driving practical. Platforms such as Waymax \cite{waymax}, Nocturne \cite{nocturne}, and GPUDrive \cite{kazemkhani2025gpudrivedatadrivenmultiagentdriving} support parallel simulation and enable millions of agent steps per second. These systems provide the computational substrate for modern self-play training, but they typically assume access to logged trajectory data or homogeneous vehicle dynamics. Our work builds on GPUDrive and extends this setting in two directions: synthetic scenario generation without expert trajectory imitation, and heterogeneous dynamics that support both passenger cars and articulated trucks within the same training framework.

\textbf{Imitation learning for behavior and interaction modeling.}
Imitation-learning approaches such as Versatile Behavior Diffusion \cite{VBD} and BehaviorGPT \cite{behaviorGPT} have set strong benchmarks for realistic traffic generation by modeling the multimodality of road-user trajectories. Their main strength is fidelity to observed data. However, because behavior remains tied to the training distribution, these methods offer limited explicit controllability and can suffer from compounding errors in closed-loop settings when rollouts leave the support of the data. Our approach instead optimizes behavior in closed loop and exposes direct controls through conditioning on goals and context variables such as speed, action range, and vehicle type, enabling targeted scenario generation while preserving realistic interaction dynamics.

\textbf{Trajectory Realism and Distributional Evaluation.}
Evaluating traffic realism remains an active area of research. The Waymo Open Sim Agents Challenge (WOSAC) \cite{waymosimagent} popularized distributional evaluation of simulated rollouts against real trajectories, complementing task-level measures such as collision rate or goal completion. Following this line of work, we evaluate realism in a learned trajectory embedding space induced by a frozen motion-forecasting model \cite{Ngiam2022SceneTransformer}, together with forecasting-style displacement metrics, to assess whether simulated behavior aligns with real highway driving beyond simple task success.

\section{Method}
\subsection{Task Definition}\label{sec:taskdef}
We formulate highway traffic generation as a Partially Observable Stochastic Game (POSG)~\cite{posg}, defined by the tuple
\[
\langle \mathcal{I}, \mathcal{S}, \{\mathcal{O}^i\}, \{\mathcal{A}^i\}, \mathcal{P}, \{r^i\}, \gamma, \mathcal{C} \rangle .
\]
Here, $\mathcal{I} = \{1, \dots, N\}$ denotes the set of agents, $\mathcal{S}$ is the global joint state space, and $\mathcal{O}^i$ and $\mathcal{A}^i$ are the observation and action spaces of agent $i$.

We define a \emph{conditioned goal-reaching} task. Each agent $i$ is assigned a Cartesian goal location $g_i \in \mathbb{R}^2$ together with an individual conditioning context vector
\[
\mathcal{C}_i = (v_{\text{goal},i}, \alpha_i, \mathcal{T}_i, \mathcal{D}_i).
\]
Here, $v_{\text{goal},i} \in [0,40]$~m/s is the target cruising speed, $\alpha_i \in [0.1,1]$ modulates the agent's longitudinal control range, $\mathcal{T}_i \in \{\text{Car}, \text{Truck}\}$ specifies vehicle type, and $\mathcal{D}_i = (\ell_i, w_i, \ell_i^{\text{tr}}, w_i^{\text{tr}})$ specifies vehicle dimensions, including corresponding trailer dimensions for articulated vehicles. Trailer terms are zero for passenger cars.

An episode is considered successful for agent $i$ if the agent reaches a designated target region around $g_i$ without collision over a finite horizon $T$. Target speed and yaw alignment are treated as soft objectives through reward shaping, while $\alpha_i$ defines the agent's control envelope rather than an additional success criterion.

The transition function $\mathcal{P}(s' \mid s, \mathbf{a})$ is determined by the underlying vehicle-dependent kinematics induced by $\mathcal{T}_i$ and $\mathcal{D}_i$. 
Because each agent's return depends directly on the actions of the other agents, the environment is inherently non-stationary from any single-agent perspective, motivating self-play for learning decentralized policies that induce coherent multi-agent traffic behavior.

\subsection{Synthetic Scenario Generation}
We train \RLPOLICY{} entirely in procedurally generated highway scenes to maximize coverage over traffic compositions, kinematics, and interaction patterns without relying on expert trajectory logs. The generator is controlled by seven parameters: the lane-change ratio $P_{lc}$, truck proportion $P_{\text{truck}}$, agent-count bounds $N_{\min}$ and $N_{\max}$, path-distance bounds $D_{\min}$ and $D_{\max}$, and a lane-change budget $K$.

\begin{figure}[t]
  \centering
  \includegraphics[width=\columnwidth]{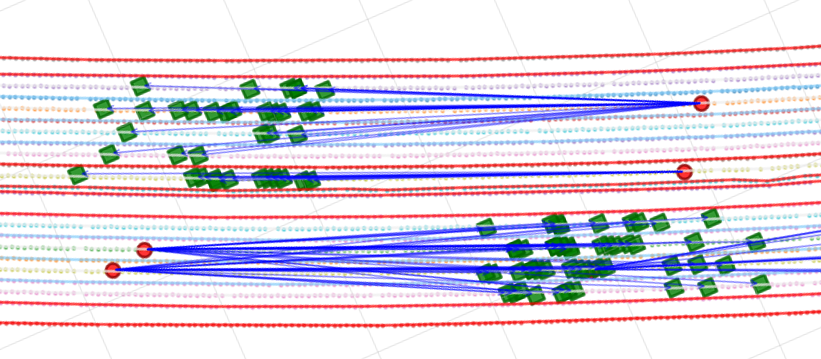}
  \caption{\textbf{Example start--goal pairs sampled from the offline lane-graph pool.} Red nodes denote start points, green nodes denote sampled goal points, and blue lines connect each start to its candidate goals. The map overlay shows road boundaries in red, lane boundaries in light blue, and lane centers in gray.}
  \label{fig:sample_start_goal}
\end{figure}

Offline, for each map we construct a reusable tuple $(\text{map}, \text{pool})$, where the pool contains candidate start--goal pairs. We first upsample the road geometry into a lane graph $G$ with nodes spaced at 1~m intervals and edges encoding lane continuity and left/right adjacency. For each sampled node, we run a bounded breadth-first search over longitudinal and lateral transitions to enumerate reachable endpoints whose path length lies in $[D_{\min}, D_{\max}]$. Each frontier state tracks a signed lane-change count $c$, where left and right transitions update $c$ by $-1$ and $+1$, respectively, and only states with $|c| \le K$ are retained. Candidate endpoints are then partitioned into same-lane ($c=0$) and lane-change ($c \neq 0$) sets, from which we sample goals to match the target lane-change ratio $P_{lc}$. Figure~\ref{fig:sample_start_goal} shows example start--goal pairs from this offline pool.

Online, we instantiate multiple worlds from each $(\text{map}, \text{pool})$ tuple and sample scenarios independently within each world.
We begin by sampling the number of agents as $N \sim \mathcal{U}[N_{\min}, N_{\max}]$. A fraction $P_{\text{truck}}$ of the sampled agents are then designated as trucks and assigned truck-specific dynamics, while the remaining agents use car dynamics. We also randomize vehicle geometry on a per-agent basis. Finally, subject to collision checking, we sample each agent's initial state and goal from the offline pool.

To set kinematics, we compute a base speed $v_{\text{base}}$ from the episode horizon and the average path length $(D_{\min}+D_{\max})/2$, including safety buffers. Each agent then receives an initial speed
\[
v_{\text{init},i} = v_{\text{base}} (1 + \epsilon_{\text{init},i}),
\]
where $\epsilon_{\text{init},i}$ is a per-agent perturbation. We further sample a goal-speed offset $\epsilon_{\text{goal},i}$ and define
\[
v_{\text{goal},i} = v_{\text{init},i} (1 + \epsilon_{\text{goal},i}),
\]
together with an action-range parameter $\alpha_i \sim \mathcal{U}(0.1,1.0)$.

This offline--online design separates map-consistent route generation from online world randomization. In practice, it yields broad coverage over lane-change structure, traffic density, vehicle type, geometry, and target behavior, while preserving control through per-agent conditioning.

\subsection{Observations, Actions, and Kinematic Models}
\textbf{Observation Space.} 
We use an ego-centric observation space composed of the ego state, the 16 nearest neighboring agents, and local road geometry. 
The ego state includes current speed, vehicle dimensions, trailer dimensions, hitch angle, a truck indicator, relative goal position and heading, and the longitudinal action-range conditioning variable $\alpha_i$. To capture short-term control history, we also include current acceleration, steering angle, relative heading, and a 3-step history of acceleration and steering. Neighbor observations encode relative position, speed, orientation, and articulated bounding-box dimensions for surrounding traffic participants. Local road topology is represented by categorized spatial points describing drivable paths and boundaries. To improve robustness, we inject Gaussian noise into partner and map observations.

\textbf{Action Space.} 
The action space is discrete and defined as the Cartesian product of bounded longitudinal and lateral inputs, specifically longitudinal jerk (m/s$^3$) and steering rate (rad/s). At each step, the policy produces a discrete action token, which is subsequently mapped to continuous commands before being passed to the simulator. The selected longitudinal command is then scaled by the agent-specific conditioning variable $\alpha_i$, allowing the same policy structure to adapt its behavior to different vehicle capabilities. In practice, this conditioning allows a single policy to operate across a wide range of control envelopes, spanning agile passenger cars as well as heavy articulated trucks.

\textbf{Kinematic Model.} 
Each agent is propagated with a standard discrete-time kinematic bicycle model~\cite{bicycle} using longitudinal jerk $j$ and steering rate $\dot{\delta}$. For articulated vehicles, we additionally update the hitch angle state $\phi$, clipped to $[-\pi/2, \pi/2]$, according to
\begin{equation}
\phi_{t+1}=\phi_t+\left(\dot{\theta}_t-\frac{v_t}{l_{\text{trailer}}}\sin\phi_t\right)\Delta t
\label{eq:hitch_angle_update}
\end{equation}
where $l_{\text{trailer}}$ is the trailer length.

\subsection{Simulation Mechanisms}\label{sec:mec}
We introduce two simulation mechanisms to improve training stability: early termination of unrecoverable states and at-fault collision attribution.

\begin{figure}[t]
  \centering
  \includegraphics[width=\linewidth]{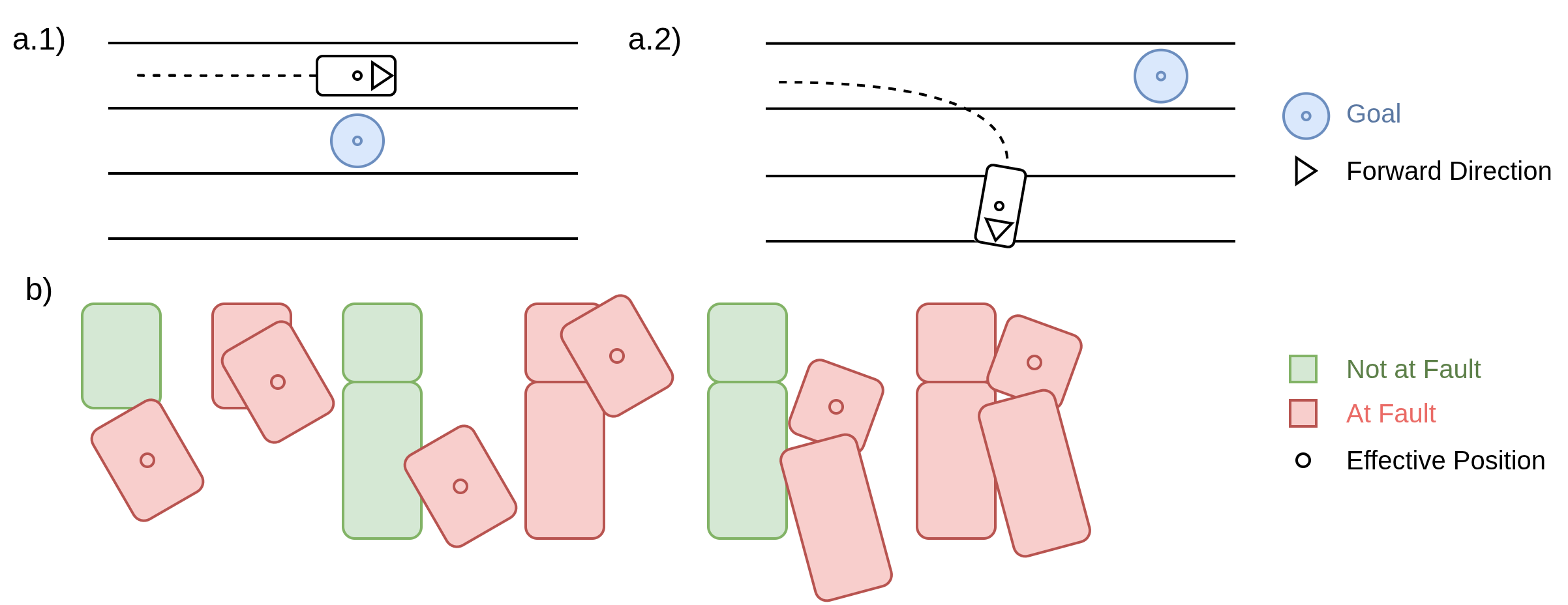}
  \caption{\textbf{Simulation mechanisms used for stabilization.} Panels (a.1) and (a.2) show unrecoverable states in which the agent's forward direction diverges from the goal direction, triggering early termination. Panel (b) illustrates collision fault attribution based on relative positions.}
  \label{fig:sim_mechanisms}
\end{figure}

We early-terminate an agent when it enters a state from which reaching its assigned goal is no longer realistically plausible. Let $g_i \in \mathbb{R}^2$ denote the goal position expressed in the agent's local frame, and let $f_i \in \mathbb{R}^2$ denote the agent's forward unit vector. Agent $A_i$ is classified as unrecoverable if $g_i^\top f_i < 0$. Intuitively, this condition identifies states in which the goal lies behind the agent, indicating that the agent is moving away from its destination. Figure~\ref{fig:sim_mechanisms} illustrates two examples of such unrecoverable states in cases (a.1) and (a.2). This criterion provides two benefits:
\begin{enumerate}
  \item \label{itm:mec-near-goal} It terminates agents traveling in an adjacent lane when the goal lies immediately to the left or right, thereby preventing unrealistic last-moment lane changes caused by extreme steering corrections near the goal.
  \item \label{itm:mec-sample-eff} During the early stages of training, it prevents agents from collecting low-value trajectories that move away from the goal, which in turn improves sample efficiency and helps stabilize training.
\end{enumerate}

As in prior self-play work~\cite{cornelisse2025buildingreliablesimdriving}, we do not resolve collisions physically and instead penalize them through rewards. Unlike prior approaches, however, we penalize only the at-fault agent. We find this attribution rule important for training stability and performance. Formally, for a colliding pair $(A_i, A_j)$, let $p_i, p_j \in \mathbb{R}^2$ denote effective global positions, and let $\ell_j$ denote the effective tractor or vehicle length of $A_j$. We assign fault to $A_i$ if
\begin{equation}
F_{i} = \mathbb{1}\!\left[(p_j - p_i)^\top f_j > \tfrac{1}{2}\ell_j\right].
\end{equation}
We define $F_j$ analogously. If neither agent is clearly behind the other (i.e., $F_i = F_j = 0$), we conservatively assign fault to both ($F_i \gets 1, F_j \gets 1$). 
Figure~\ref{fig:sim_mechanisms} (b) illustrates this collision-attribution mechanism.

\subsection{Rewards}\label{sec:rewards}
We use a mixture of sparse and dense reward terms to encourage safe, goal-consistent highway behavior while retaining useful learning signal early in training. For each agent $A_i$, the per-step reward is
\begin{equation}
\begin{aligned}
R_i = {} & R_{g,i} + R_{l,i} - R_{f,i} - R_{e,i}\\
         & - R_{t,i} - R_{a,i} - R_{s,i} + R_{p,i}.
\end{aligned}
\end{equation}
Here, $R_{g,i}$ corresponds to goal completion, $R_{l,i}$ to lane-boundary compliance, $R_{f,i}$ to collision penalty, $R_{e,i}$ to road-edge penalty, $R_{t,i}$ to early-termination penalty, $R_{a,i}$ to alignment penalty, $R_{s,i}$ to speed-deviation penalty, and $R_{p,i}$ progress reward. 
We define each term as follows:
\begin{alignat}{2}
R_{g,i} &{}= \bar{w}_{g} m_{g,i}(\rho, w_{s,i}, w_{a,i}) G_i ,\\
R_{l,i} &{}= \bar{w}_{l} (1 - L_i) ,\\
R_{f,i} &{}= \bar{w}_{f} m_f(\rho) F_i ,\\
R_{e,i} &{}= \bar{w}_{e} m_e(\rho) E_i ,\\
R_{t,i} &{}= \bar{w}_{t} m_t(\rho) T_i ,\\
R_{a,i} &{}= \bar{w}_{a} r_{\text{alignment},i} ,\\
R_{s,i} &{}= \bar{w}_{s} r_{\text{speed},i} ,\\
R_{p,i} &{}= \bar{w}_{p} m_p(\rho) r_{\text{progress},i} .
\end{alignat}
Here, $\bar{w}_{(\cdot)} > 0$ are fixed weights and $\rho \in [0,1]$ denotes curriculum progress. 
The curriculum multipliers $m_{g,i}(\rho, w_{s,i}, w_{a,i})$, $m_f(\rho)$, $m_e(\rho)$, $m_t(\rho)$, and $m_p(\rho)$ are defined in Section~\ref{sec:curriculums} as functions of curriculum progress and, where applicable, goal-quality terms. Here, $F_i$ denotes the collision indicator, $G_i$ denotes the goal-achievement indicator, $E_i$ denotes the road-edge collision indicator, and $L_i$ denotes the lane-boundary collision indicator.

For early termination, we set $T_i \gets \|g_i\|_2$ when agent $i$ is terminated under the unrecoverable-state rule in Section~\ref{sec:mec}. This penalizes goal-divergent failures more strongly when they occur far from the goal, while assigning smaller penalties to near-goal failures.

We use $w_{s,i}, w_{a,i} \in \{0.1,1\}$ to represent goal completion quality. The factor $w_{s,i}$ measures agreement with the target speed at the goal, and $w_{a,i}$ measures yaw alignment with the lane direction at the goal. These terms act as soft terminal preferences rather than hard feasibility constraints.

The progress reward $r_{\text{progress},i}$ provides dense guidance toward the goal, especially early in training:
\begin{equation}
r_{\text{progress},i} = \mathrm{clip}\!\left( \frac{d_{t-1,i} - \kappa\, d_{t,i}}{v_{t,i}}\, \psi(d_{t,i}),\, -0.5,\, 0.5 \right).
\end{equation}
Here, $d_{t,i}$ is the Euclidean distance to the goal, $\kappa > 1$ is a progress factor, and $\psi(d)$ is a distance-dependent decay term. The reward is normalized by speed so that faster agents are not favored purely because they cover more distance, and it is clipped for stability.

The alignment term $r_{\text{alignment},i}$ captures the increased pull to perform lane changes as agent $i$ moves closer to its goal:
\begin{equation}
r_{\text{alignment},i} = \Delta \theta_i \cdot \mathrm{clip}\!\left(1 - \frac{d_{t,i}}{v_{t,i} \bar{T_{\text{ramp}}}},\, 0,\, 1\right) .
\end{equation}
Here, $\Delta\theta_i$ is the yaw-error term for agent $i$ and $\bar{T_{\text{ramp}}}$ is the alignment ramp horizon. 
This term increases as the agent approaches its goal, encouraging timely lane changes and reducing late, aggressive corrections. The ramp depends on an estimate of time-to-go, rather than raw distance, so that slower agents are not permitted larger heading errors at the same spatial distance.

The speed term $r_{\text{speed},i}$ encourages the agent to track its assigned target speed $v_{\text{goal},i}$ throughout the rollout:
\begin{equation}
r_{\text{speed},i} = |v_{t,i} - v_{\text{goal},i}|.
\end{equation}

\subsection{Curricula} \label{sec:curriculums}
We use two coupled curricula: a reward curriculum and a scenario curriculum.

For the reward curriculum, we introduce a curriculum progress variable $\rho \in [0,1]$ and define:
\begin{equation}
m_{g,i}(\rho, w_{s,i}, w_{a,i}) = 1 - \rho \left(1 - w_{s,i} w_{a,i}\right)
\end{equation}
\begin{equation}
m_f(\rho)=m_e(\rho)=m_t(\rho)=1 + (\lambda - 1)\rho
\end{equation}
\begin{equation}
m_p(\rho)=1-\rho
\end{equation}
Here, $w_{s,i} w_{a,i}$ is the goal-reward scaling term, and $\lambda > 1$ is the terminal curriculum multiplier that controls how strongly collision and off-road penalties are increased by the end of the curriculum. Thus, the goal-achievement multiplier is annealed, collision/off-road multipliers are ramped up, and the progress multiplier decays to $0$ at the terminal curriculum stage. 
Intuitively, this schedule gradually shifts optimization from dense guidance toward stricter emphasis on safety, alignment, and terminal behavior. Early in training, the progress reward helps agents discover goal-directed behavior; later, collision and off-road penalties become more dominant.

For the scenario curriculum, we increase the lane-change ratio $P_{lc}$ and shift the agent-count distribution toward denser scenes at the end of training ($\rho = 1$). This late-stage curriculum exposes the policy to more congested, interaction-heavy traffic after it has already learned basic driving structure, which improves robustness and stability in challenging highway regimes.

\begin{figure}[t]
  \centering
  \includegraphics[width=\linewidth]{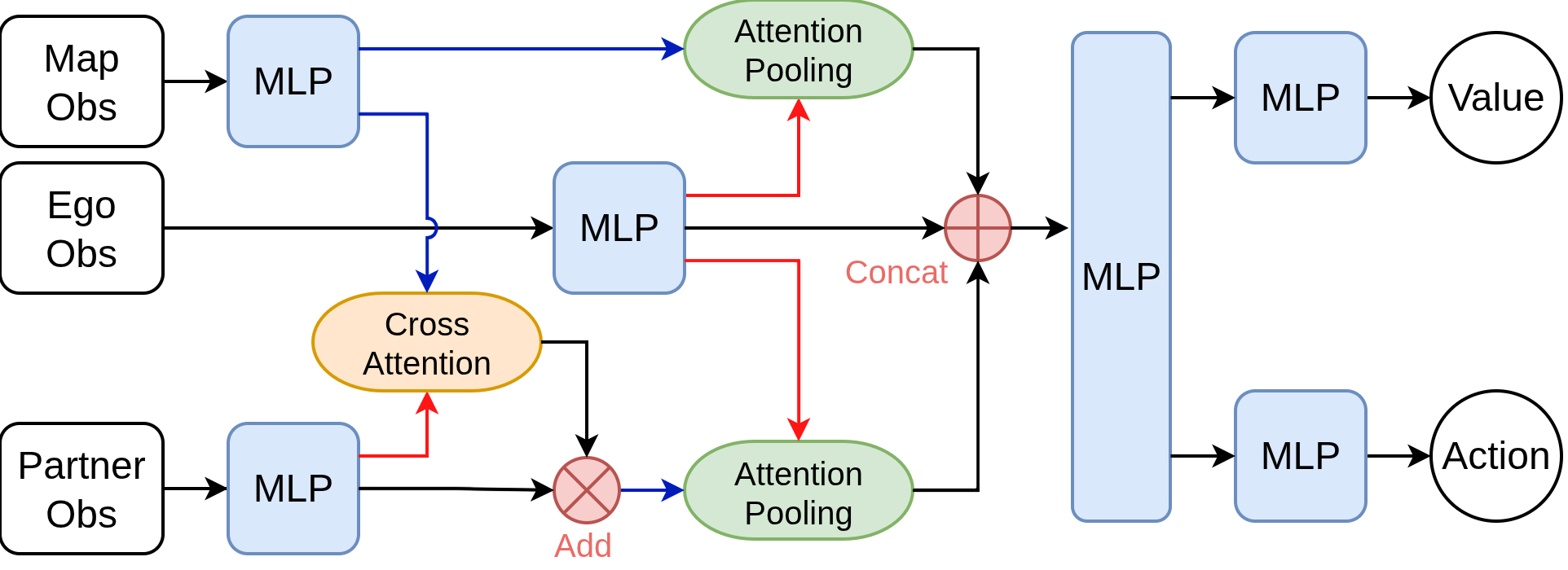}
  \caption{\textbf{Model architecture of \RLPOLICY.} Ego, partner-agent, and road-element features are encoded and fused via cross-attention and attention pooling. Red lines denote queries, and blue lines denote keys and values.}
  \label{fig:network}
\end{figure}

\subsection{Model}
We parameterize \RLPOLICY{} with an MLP-based architecture augmented with cross-attention~\cite{Vaswani2017Attention} and attention pooling~\cite{Lee2019SetTransformer}, as shown in Figure~\ref{fig:network}. Ego, partner-agent, and road-element features are first encoded separately. We then apply cross-attention between partner and road features to capture interactions between neighboring agents and local topology, followed by attention pooling to aggregate variable-sized inputs while preserving permutation invariance. In addition to the encoded observation, the policy receives the conditioning variables $(v_{\text{goal},i}, \alpha_i, \mathcal{T}_i, \mathcal{D}_i)$ as explicit inputs, which are concatenated with the ego feature embedding before the policy and value heads. The resulting representation is passed to separate policy and value heads.

\begin{table*}[!tbh]
\centering
\begin{minipage}[t]{0.495\textwidth}
\centering
\small
\setlength{\tabcolsep}{4.5pt}
\begin{tabular}{cccccc}
\toprule
Tuples & $N^{\mathrm{pre}}$ & $N^{\mathrm{post}}$ & $v_{\text{init}}$ (m/s) & $P_{lc}^{\mathrm{pre}}$ & $P_{lc}^{\mathrm{post}}$ \\
\midrule
22 & $176 \pm 80$ & $200 \pm 56$ & $6.7 \pm 2.7$ & 0.4 & 0.5 \\
22 & $80 \pm 16$ & $120 \pm 24$ & $15.0 \pm 6.0$ & 0.4 & 0.5 \\
22 & $48 \pm 16$ & $80 \pm 16$ & $23.3 \pm 9.3$ & 0.4 & 0.5 \\
11 & $16 \pm 16$ & $32 \pm 32$ & $31.6 \pm 12.6$ & 0.4 & 0.6 \\
\bottomrule
\end{tabular}
\caption{Training scenario statistics for the offline--online synthetic generator. Each row corresponds to a group of $(\text{map}, \text{pool})$ tuples with different traffic-density and speed regimes. $N^{\mathrm{pre}}$ and $N^{\mathrm{post}}$ denote the agent-count distributions before and after the curriculum switch, $P_{lc}^{\mathrm{pre}}$ and $P_{lc}^{\mathrm{post}}$ denote the corresponding lane-change proportions, and $K=3$ is fixed across all tuples.}
\label{tab:scenario_stats}
\end{minipage}
\hfill
\begin{minipage}[t]{0.495\textwidth}
\centering
\small
\begin{tabular}{@{}l c|l c|l c@{}}
\toprule
Symbol & Value & Symbol & Value & Symbol & Value \\
\midrule
$\bar{w}_{g}$ & 3.3 & $\bar{w}_{t}$ & 8.3 & $\bar{w}_{f}$ & 1.0 \\
$\bar{w}_{e}$ & 0.9 & $\bar{w}_{l}$ & 0.005 & $\bar{w}_{p}$ & 0.25 \\
$\bar{w}_{a}$ & 0.09 & $\bar{w}_{s}$ & 0.07 & $\lambda$ & 2.69 \\
$\kappa$ & 1.01 & $d_{\min}$ & 50m & $d_{\max}$ & 130m \\
$\bar{T_{\text{ramp}}}$ & 5s &  &  &  &  \\
\bottomrule
\end{tabular}
\caption{Constant reward and curriculum parameters used throughout training. The listed symbols correspond to the reward terms defined in Section~\ref{sec:rewards} and the curriculum schedules described in Section~\ref{sec:curriculums}. In particular, $d_{\min}$ and $d_{\max}$ denote the lower and upper path-distance bounds used in scenario generation.}
\label{tab:reward_constants}
\end{minipage}
\end{table*}

\begin{table*}[t]
\centering
\small
\setlength{\tabcolsep}{10.0pt}
\begin{tabular}{lccccccccc}
\toprule
Method & \shortstack{$GR (\%)$ $\uparrow$} & \shortstack{$CR_a (\%)$ $\downarrow$} & \shortstack{$CR_r (\%)$ $\downarrow$} & \shortstack{$SR (\%)$ $\uparrow$} & \shortstack{ADE$_{5s} (m)$ $\downarrow$} & \shortstack{FDE$_{5s} (m)$ $\downarrow$} \\
\midrule
\textbf{\RLPOLICY} & $\mathbf{96.8 \pm 5.3}$ & $\mathbf{0.7 \pm 2.6}$ & $1.2 \pm 3.1$ & $\mathbf{96.3 \pm 5.8}$ & $2.44 \pm 0.25$ & $5.25 \pm 1.80$\\
\quad$-$ DClamp & $93.9 \pm 7.8$ & $2.9 \pm 7.8$ & $1.8 \pm 4.0 $ & $91.9 \pm 8.8$ & $\mathbf{2.12 \pm 0.51}$ & $\mathbf{4.97 \pm 1.45}$\\
\quad$-$ sample reweighting & $89.56 \pm 9.4$ & $3.7 \pm 6.3$ & $ \mathbf{1.1 \pm 2.6} $ & $87.3 \pm 10.8$ & $2.54 \pm 0.76$ & $7.00 \pm 2.03$\\
\quad$-$ curriculum & $85.5 \pm 10.2$ & $2.3 \pm 4.9$ & $7.8 \pm 8.3 $ & $83.9 \pm 11.0$ & $2.86 \pm 0.60$ & $8.08 \pm 2.05$\\
\midrule
Cornelisse et al.~\cite{cornelisse2025buildingreliablesimdriving} & $35.3 \pm 38.0$ & $19.5 \pm 18.0$ & $24.7 \pm 21.7$ & $26.6 \pm 30.1$ &$6.57 \pm 2.62$ & $12.07 \pm 4.09$ \\
\quad$+$ same model & $63.4 \pm 21.0$ & $31.2 \pm 21.4$ & $29.5 \pm 26.8$ & $43.7 \pm19.1$ & $7.62 \pm 1.96$ & $17.21 \pm 5.77$  \\
\bottomrule
\end{tabular}%
\caption{Zero-shot evaluation on 512 exiD scenarios and ablation comparison across policy variants. Arrows indicate whether higher or lower is better. $GR$ denotes goal-reaching rate, $CR_a$ agent-collision rate, $CR_r$ road-edge collision rate, and $SR$ success rate, defined as reaching the goal without causing a collision with another agent. ADE and FDE are computed over a 5~s horizon. The full \RLPOLICY{} configuration achieves the strongest overall performance despite minor regression in ADE/FDE/$CR_r$}
\label{tab:main_ablation}
\end{table*}

\subsection{Training}
Highway self-play is less stable than the lower-speed urban settings considered in prior work because speed, density, and interaction complexity vary widely across scenarios. The same map can generate sparse, fast traffic or dense, slow traffic, and a single policy must learn across this full distribution. We therefore use three complementary design choices to improve training stability: DClamp-PPO for stable policy updates under coupled curriculum shifts, inverse-agent-count reweighting to balance gradient contributions across traffic densities, and action regularization toward smooth, near-zero-centered control.

\textbf{Training Algorithms.} 
We use DClamp-PPO~\cite{dclamp}, a variant of PPO~\cite{ppo}, to reduce instability during curriculum transitions. As curriculum progress $\rho$ increases, both the reward landscape and scenario distribution shift: dense progress guidance is reduced, terminal penalties become stronger, and later-stage scenarios become denser and more interactive. These coupled changes can cause standard PPO to overshoot. DClamp-PPO imposes a tighter effective trust region during such transitions, leading to smoother adaptation and fewer late-stage collapses.

\textbf{Sample Reweighting.} 
Worlds with many agents produce more trajectories per update than sparse worlds. Without correction, dense worlds dominate the gradient and bias training toward slow, crowded traffic. To balance contributions across traffic regimes, we weight each world's loss inversely by its agent count:
\begin{equation}
\mathcal{L}_{\pi}^{\mathrm{rw}} = \sum_{w \in \mathcal{W}} \frac{1}{N_w} \sum_{i=1}^{N_w} \mathcal{L}_{\pi}^{(w,i)}.
\label{eq:sample_reweighting}
\end{equation}
Here, $\mathcal{W}$ is the set of sampled worlds in an update, $N_w$ is the number of agents in world $w$, and $\mathcal{L}_{\pi}^{(w,i)}$ is the policy loss for agent $i$ in world $w$. We apply the same reweighting to the value loss. This normalization makes each world contribute more evenly regardless of density and reduces the variance amplification induced by highly interactive dense scenes.

\textbf{Action Regularization.}
We replace the standard PPO entropy bonus with a KL regularizer toward a discrete action prior induced by a zero-mean Gaussian over the jerk--steering lattice. The prior is parameterized as $\mathcal{N}(0, \Sigma_0)$, centered at zero jerk and zero steering rate. Concretely, we optimize:
\begin{equation}
\mathcal{L}_{\pi}^{\mathrm{total}}
=
\mathcal{L}_{\pi}^{\mathrm{rw}}
-
\lambda_{\mathrm{KL}}
\,\mathbb{E}_{s\sim\mathcal{D}}
\!\left[
D_{\mathrm{KL}}
\!\left(
\pi_{\theta}(\cdot\mid s)\,\|\,\mathcal{N}(0, \Sigma_0)
\right)
\right]
\label{eq:kl_action_reg}
\end{equation}
where $\lambda_{\mathrm{KL}} > 0$ controls the regularization strength and $\Sigma_0 \in \mathbb{R}^2$ sets the spread of the prior for steering rate and jerk invidually. This regularizer favors moderate, smoother controls while encouraging exploration.

\section{Experiments and Results}
We train a 670K-parameter policy from scratch on 22 highway maps in the United States, each spanning approximately 500~m. From these maps, we construct 77 distinct $(\text{map}, \text{pool})$ tuples covering a range of speed regimes, traffic densities, and lane-change proportions, and instantiate 300 parallel worlds in GPUDrive~\cite{kazemkhani2025gpudrivedatadrivenmultiagentdriving}. Key scenario statistics are summarized in Table~\ref{tab:scenario_stats}. Training runs for 3 billion environment steps and takes approximately 80 hours on two NVIDIA A6000 GPUs. Table~\ref{tab:reward_constants} reports the constant reward and curriculum parameters used throughout training.

\subsection{Zero-shot Real-world Scenario Evaluation}
We first evaluate zero-shot transfer on exiD~\cite{exid}, a real-world highway dataset with dense multi-agent interactions. We randomly sample 512 scenarios of length 10~s that contain at least one lane-change event. Agents are initialized at the start of each scenario and rolled out until the end, with action-range conditioning $\alpha_i$ sampled at random. All reported metrics are aggregated over the 512 scenarios and shown as mean $\pm$ standard deviation.

All policies are trained on the same synthetic training distribution using Optuna-optimized~\cite{optuna} hyperparameters, and are then evaluated zero-shot on exiD without fine-tuning. Table~\ref{tab:main_ablation} compares \RLPOLICY{} against a prior self-play baseline~\cite{cornelisse2025buildingreliablesimdriving}, together with ablations of both approaches. The -DClamp PHASE variant uses a standard PPO~\cite{ppo} instead of DClamp-PPO~\cite{dclamp}.

\RLPOLICY{} substantially outperforms the prior self-play baseline across all task-level and trajectory-level metrics. In particular, it achieves a success rate of 96.3\%, compared with 26.6\% for the prior baseline, while reducing ADE/FDE from 6.57/12.07~m to 2.44/5.25~m. These gains indicate that the proposed conditioning, synthetic scenario generation, and stabilization design materially improve transfer to previously unseen real highway scenes.

\begin{table*}[t]
\centering
\small
\setlength{\tabcolsep}{4pt}
\begin{tabular}{lcc ccccc cc}
\toprule
\multirow{2}{*}{Method}
& \multirow{2}{*}{Fréchet $\downarrow$}
& \multirow{2}{*}{Energy $\downarrow$}
& \multicolumn{5}{c}{MMD $\downarrow$}
& \multirow{2}{*}{Precision $\uparrow$}
& \multirow{2}{*}{Recall $\uparrow$} \\
\cmidrule(lr){4-8}
& &
& $\gamma=7{\times}10^{-5}$
& $\gamma=10^{-4}$
& $\gamma=5{\times}10^{-4}$
& $\gamma=10^{-3}$
& $\gamma=5{\times}10^{-3}$
& & \\
\midrule
IDM \cite{idm}
& 6.3501
& 0.0233
& 1.37e-4
& 1.67e-4
& 2.46e-4
& 2.07e-4
& 5.40e-5
& 0.9684
& 0.9999 \\
\RLPOLICY
& \textbf{5.5199}
& \textbf{0.0186}
& \textbf{1.07e-4}
& \textbf{1.34e-4}
& \textbf{1.98e-4}
& \textbf{1.67e-4}
& \textbf{4.00e-5}
& \textbf{0.9690}
& \textbf{0.9999} \\
\bottomrule
\end{tabular}
\caption{Distributional similarity between simulated rollouts and real highway trajectories in a learned embedding space. Lower is better for Fréchet distance, energy distance, and MMD; higher is better for precision and recall. Across all reported distributional metrics, \RLPOLICY{} is closer to the real trajectory distribution than the IDM baseline.}
\label{tab:distributional}
\end{table*}

\subsection{Distributional Evaluation}
Task completion alone does not determine whether a simulator reproduces realistic traffic behavior. We therefore evaluate whether the rollouts generated by \RLPOLICY{} match the statistical structure of real highway trajectories in a learned latent space. Following the evaluation perspective introduced in the Waymo Sim Agents Challenge~\cite{waymosimagent}, we compare simulated and real trajectories using distributional metrics rather than only task-level outcomes.

Each multi-agent trajectory sequence is embedded using the encoder of a state-of-the-art motion forecasting model~\cite{Ngiam2022SceneTransformer}, and we compare the resulting simulated and real trajectory distributions in the embedding space using Fréchet distance, energy distance, and Maximum Mean Discrepancy (MMD) with an RBF kernel. Our evaluation uses 33,398 trajectories from proprietary real highway driving logs together with corresponding simulator rollouts, and compares \RLPOLICY{} against a classical rule-based microscopic traffic model based on the Intelligent Driver Model (IDM)~\cite{idm}; Table~\ref{tab:distributional} reports the results.

Across all reported metrics, \RLPOLICY{} yields closer agreement with real trajectory distributions than IDM. In particular, Fréchet distance decreases from 6.3501 to 5.5199 and energy distance decreases from 0.0233 to 0.0186. MMD is also consistently lower across all tested RBF bandwidths, indicating that the improvement is not tied to a specific kernel scale. Precision and recall in the embedding space further suggest that both methods cover a broad portion of the real trajectory manifold, while \RLPOLICY{} produces slightly fewer out-of-distribution behaviors.

\begin{figure}[t]
    \centering
    \includegraphics[width=\columnwidth]{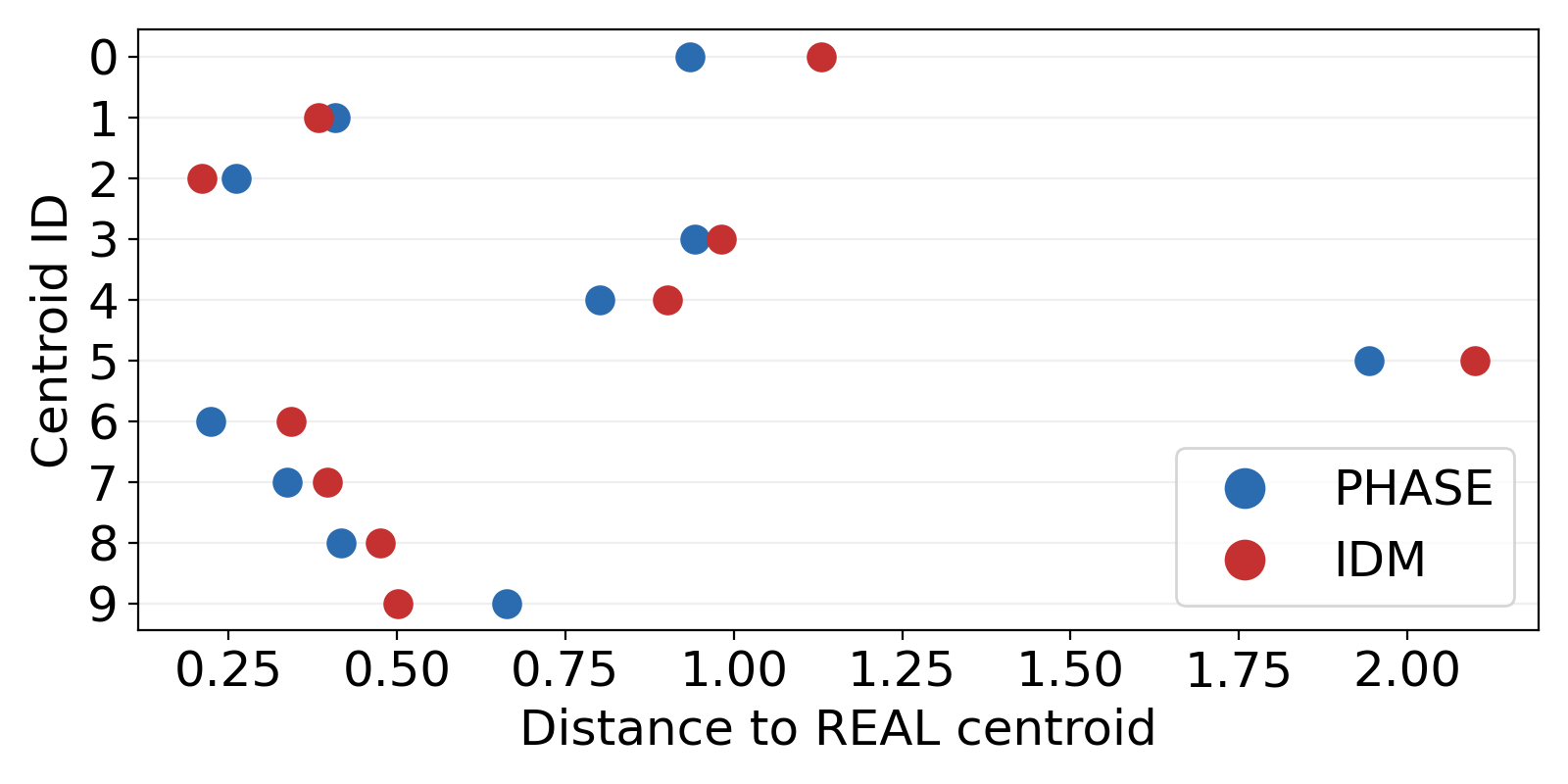}
    \caption{\textbf{Cluster-wise embedding distance to real trajectories.} Each row corresponds to a cluster in the trajectory embedding space. Blue markers denote the distance between the centroid of rollouts generated by PHASE and the centroid of real trajectories for the same cluster, while red markers denote the corresponding distance for IDM rollouts. Smaller values indicate closer alignment with the real-data distribution. Across most clusters, PHASE produces centroid embeddings that lie closer to the real trajectory centroids than IDM, indicating improved alignment of simulated multi-agent behavior with real highway dynamics.}
    \label{fig:pca2d}
\end{figure}

Quantitatively, we compare behavior-cluster centroids in the learned embedding space instead of the full embedding cloud. For each cluster, we measure the distance from the real-trajectory centroid to the IDM and \RLPOLICY{} centroids (Figure~\ref{fig:pca2d}). Lower distances indicate better alignment with real data; \RLPOLICY{} is closer than IDM in most clusters, indicating more realistic behavior across modes.

\section{Conclusion and Future Work}
We introduced \textbf{\RLPOLICY}, a context-aware self-play framework for controllable and realistic highway traffic simulation. A single policy controls heterogeneous agents across a 0--40~m/s range, combining offline--online scenario generation, curriculum learning, and stable self-play. On exiD, \RLPOLICY{} transfers zero-shot and substantially outperforms prior self-play baselines on success and collision metrics. Distributional evaluation in a learned trajectory embedding space also shows closer alignment to real highway behavior than a classical rule-based baseline. These results suggest conditioned self-play is a practical, scalable route to realistic highway simulation for autonomous driving.

Several directions remain for future work. First, scaling training to larger models and longer horizons may further improve robustness and coverage of rare interactions. Second, richer world models, including more diverse traffic participants, could broaden the scope of the work beyond the agent-agent interactions studied here. Finally, integrating perception and control within the same framework may enable a closer connection between simulation-agent and real-world driving.

\section{Acknowledgements}
We thank Mohammadhussein Rafieisakhaei, Joseph A. Vincent, Aman Agarwal, and Mingqing Yuan for their valuable input, technical advice, and thoughtful discussions, and their contributions to the evaluation pipelines. We also gratefully acknowledge the support and guidance from PlusAI leadership, Amit Kumar, Tim Daly, and Hao Zheng.

{
    \small
    \bibliographystyle{ieeenat_fullname}
    \bibliography{main}
}


\end{document}